\documentclass[10pt,twocolumn,letterpaper]{article}

\usepackage{cvpr}              

\usepackage[dvipsnames]{xcolor}

\newcommand{\parhead}[1]{{\noindent\textbf{#1}}}

\usepackage{booktabs}       
\usepackage{multirow}
\usepackage{enumitem} 
\usepackage{colortbl} 

\definecolor{morandiblue}{RGB}{99, 137, 187}

\newcommand\blfootnote[1]{%
  \begingroup
  \renewcommand\thefootnote{}\footnote{#1}%
  \addtocounter{footnote}{-1}%
  \endgroup
}

\definecolor{cvprblue}{rgb}{0.21,0.49,0.74}
\usepackage[pagebackref,breaklinks,colorlinks,citecolor=cvprblue]{hyperref}

\def\paperID{10741} 
\def\confName{CVPR}
\def\confYear{2024}

\title{Multi-view Aggregation Network for Dichotomous Image Segmentation}

\author{Qian Yu$^\dagger$, Xiaoqi Zhao$^\dagger$, Youwei Pang$^\dagger$\\
  {\tt\small\{ms.yuqian, zxq, lartpang\}@mail.dlut.edu.cn}\\
  Lihe Zhang$^{*}$, Huchuan Lu\\
  {\tt\small\{zhanglihe, lhchuan\}@dlut.edu.cn}\\
  Dalian University of Technology
}

\begin{document}

\maketitle
\blfootnote{$\dagger$ Equal contribution.}
\blfootnote{$*$ Corresponding author.}

\begin{abstract}
  Dichotomous Image Segmentation (DIS) has recently emerged towards high-precision object segmentation from high-resolution natural images.
  When designing an effective DIS model, the main challenge is how to balance the semantic dispersion of high-resolution targets in the small receptive field and the loss of high-precision details in the large receptive field.
  Existing methods rely on tedious multiple encoder-decoder streams and stages to gradually complete the global localization and local refinement.
  Human visual system captures regions of interest by observing them from multiple views. Inspired by it, we model DIS as a multi-view object perception problem and provide a parsimonious multi-view aggregation network (MVANet), which unifies the feature fusion of the distant view and close-up view into a single stream with one encoder-decoder structure.
  With the help of the proposed multi-view complementary localization and refinement modules, our approach established long-range, profound visual interactions across multiple views, allowing the features of the detailed close-up view to focus on highly slender structures. 
  Experiments on the popular DIS-5K dataset show that our MVANet significantly outperforms state-of-the-art methods in both accuracy and speed. The source code and datasets will be publicly available at \href{https://github.com/qianyu-dlut/MVANet}{MVANet}.
\end{abstract}

\newcommand{\mymethod}{{MVANet}\xspace}

\section{Introduction}
\label{sec:intro}

High-accuracy dichotomous image segmentation (DIS)~\cite{qin2022highly} aims to accurately identify category-agnostic foreground objects within natural scenes, which is fundamental for a wide range of scene understanding applications, including
AR/VR applications~\cite{tian2022kine,qin2101boundary}, image editing~\cite{goferman2011context}, and 3D shape reconstruction~\cite{liu2021fully}.
Different from existing segmentation tasks, DIS focuses on challenging high-resolution (HR) fine-grained object segmentation. The segmentation scope encompasses a wide range of content with varying structural complexities, regardless of their characteristics.
When confronted with the task of accurately segmenting HR objects, two primary challenges arise:
1)~\textbf{The higher demand for segmentation capability.}
Due to a larger amount of intricate details in high-accuracy HR images, accurately segmenting those objects of interest requires a more complex processing pipeline and more powerful feature modelling.
And when dealing with occlusion interference, complex lighting conditions, and variable object poses, the processing of HR data also requires better adaptability and robustness compared to low-resolution (LR) data.
2)~\textbf{The more need for processing efficiency.}
The much larger size of HR images can result in slower processing speeds and more memory constraints.
This restriction hinders the further application of existing approaches to real-world scenarios such as autonomous driving~\cite{mutsch2023model,klingner2022choice} or real-time video processing~\cite{chang2022strpm,lin2021real}.
As a result, this field has higher expectations for inference efficiency in addition to ensuring algorithmic effectiveness.

\begin{figure}[t]
  \centering
  \includegraphics[width=\linewidth]{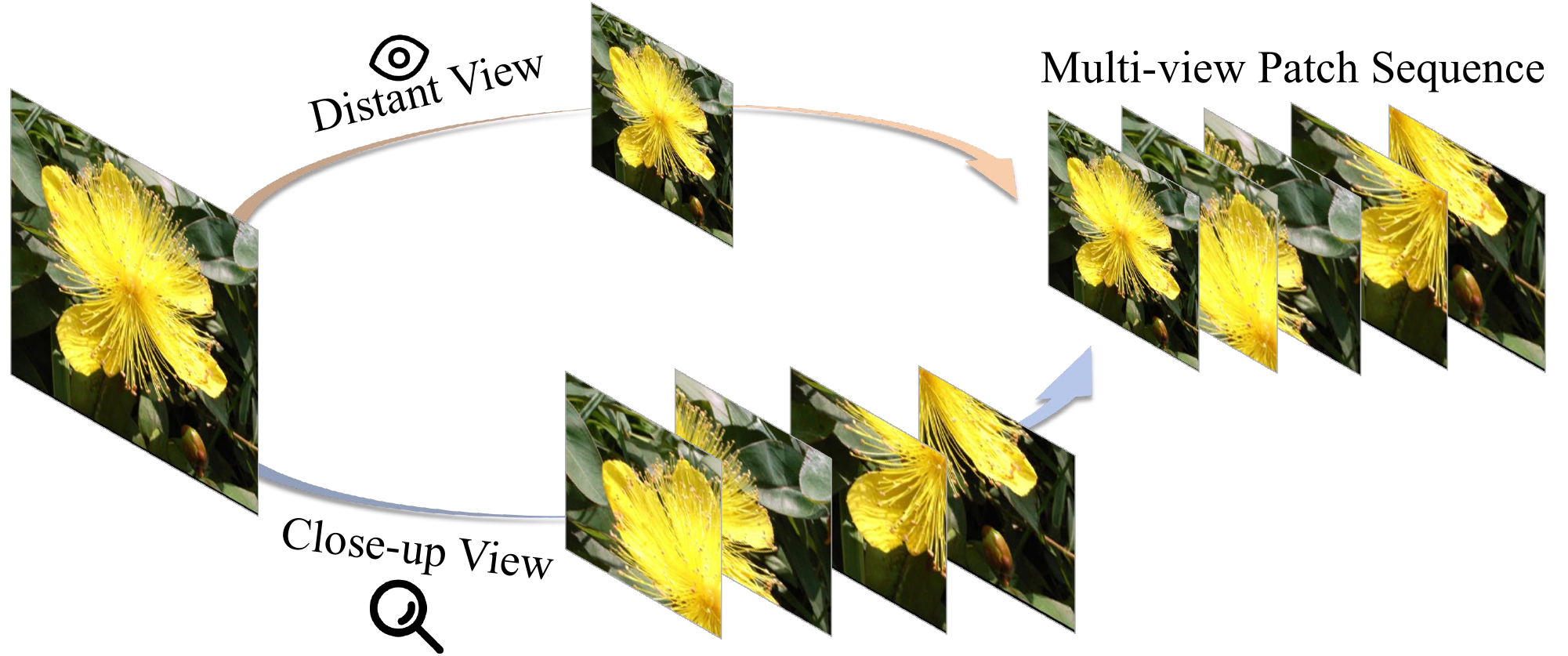}
  \caption{Process of decomposing high-resolution image into multi-view patch sequence.}
  \label{fig:multi-view}
\end{figure}
Many efforts~\cite{qin2022highly,zhoudichotomous,pei2023unite} have been made to tailor for the DIS task.
Despite existing methods have demonstrated impressive performance, their reliance on CNN may pose limitations when tackling the HR image.
It is because the increase in input resolution will result in a relatively small receptive field, which subsequently hinder the network's capacity to capture essential global semantics for the DIS task.
Recently, with the introduction of the transformer~\cite{liu2021swin,ViT} with the global information propagation capability, the transformer-based methods~\cite{xie2022pyramid,kim2022revisiting} have shown better prediction performance.
However, features are extracted with global receptive field in these transformer-based methods, but they may not handle fine-grained local details as good as CNNs, which may be detrimental in high-precision segmentation tasks.
Moreover, their multiple scale-independent models will increases the complexity of the feature pipeline and the redundancy of the model structure.
Given that the input HR image itself contains all the information in the LR image, the multi-resolution inputs in these methods may lead to repetitive computation and information redundancy.

The core of solving the aforementioned issues is to design a parallel unified framework that can be compatible with global and local cues to avoid cascading forms of feature/model reuse.
Inspired by the pattern of capturing high information content from images in the human visual system, we split the high-resolution input images from the original view into the distant view images with global information and close-up view images with local details.
Thus, they can constitute a set of complementary multi-view low-resolution input patches, as shown in \cref{fig:multi-view}.
In this paper, we make the attempt to address the HR image segmentation task by modeling it as a multi-view segmentation task.
First, we design a parsimonious multi-view aggregation network (MVANet), which obtains global semantics and local features in parallel according to the characteristics of different patches.
Such a design avoids the additional challenges caused by the hybridization of features in previous approaches.
Second, we separately propose the novel multi-view complementary localization module (MCLM) and the multi-view complementary refinement module (MCRM).
The MCLM incorporates our specially designed cross-attention mechanism driven by the global tokens and reverse attention mechanism, to enhance object localization and mitigating the local semantic gap between different patches.
The MCRM aims to achieve a detailed depiction of the localized object, which is dominated by the local tokens, which is achieved through cross-attention mechanism with modeled multi-sensory global tokens.
Subsequently, the enhanced local tokens are then used to refine the details in the global feature.
Through the two-step process, we achieve a comprehensive representation of the scene, enabling effective object segmentation that takes into account both the overall context and the intricate details.
Finally, we fuse all the patch output through a simple view rearrangement module and produce a highly accurate high-resolution prediction.

Our main contributions can be summarized as follows:
\begin{itemize}
  \item The traditional single-view high-resolution image processing mode is upgraded to a multi-view processing mode based on multi-view learning.

  \item We propose the multi-view aggregation network (MVANet), which is the first single stream and single stage framework for the dichotomous image segmentation.

  \item Two efficient transformer-based multi-view complementary localization and refinement modules are proposed to jointly capturing the localization and restoring the boundary details of the targets.

  \item \mymethod achieves state-of-the-art performance in terms of almost all metrics on the DIS benchmark dataset, while being twice as fast as the second-best method in terms of inference speed, demonstrating the superiority of our multi-view scheme.
\end{itemize}

\section{Related works}

\subsection{Dichotomous Image Segmentation}

Dichotomous image segmentation (DIS) is formulated as a category-agnostic task defined on non-conflicting annotations for accurately segmenting objects with various structural complexities, regardless of their characteristics.
What sets it apart from classic segmentation tasks is the demand for highly precise object delineation, even down to the internal details of objects. Additionally, it addresses a broader range of objects, including salient\cite{pang2020multi,zhao2020suppress}, camouflaged\cite{fan2020camouflaged,le2019anabranch}, meticulous\cite{liew2021deep,yang2020meticulous}, etc.
Many efforts have been made to tailor for DIS, the first solution, IS-Net~\cite{qin2022highly}, tackles the DIS task by employing U\(^2\)Net as backbone and leveraging the intermediate supervision strategy.
PF-DIS~\cite{zhoudichotomous} is the first to leverage frequency priors to identify fine-grained object boundaries in DIS.
Instead of using a general encoder-decoder architecture, UDUN~\cite{pei2023unite} proposes a unite-divide-unite scheme to disentangle the trunk and structure segmentation for high-accuracy DIS.
Although these works have achieved good performance, their reliance on CNN may pose limitations when tackling HR, high-accuracy tasks.
It is because the increase in input resolution will result in a relatively small receptive field, which subsequently hinders the capacity of deep networks to capture essential global semantics necessary for the DIS task.
Recently, Xie~\etal~\cite{kim2022revisiting} have proposed a novel architecture which enables to merge multiple results regardless of the size of the input. It is constructed to be trained with task-specified LR or HR inputs and generate HR output with a multi-resolution pyramid blending at the testing stage.
However, the aforementioned methods usually relay multiple stages/streams to aggregation the global and local features, which will introduce additional drawbacks such as large parameters, low efficiency, and difficulty in optimization.
In this paper, we focus on providing a parsimonious single stream and single stage baseline for the DIS task.

\begin{figure*}[t]
  \centering
  \includegraphics[width=\linewidth]{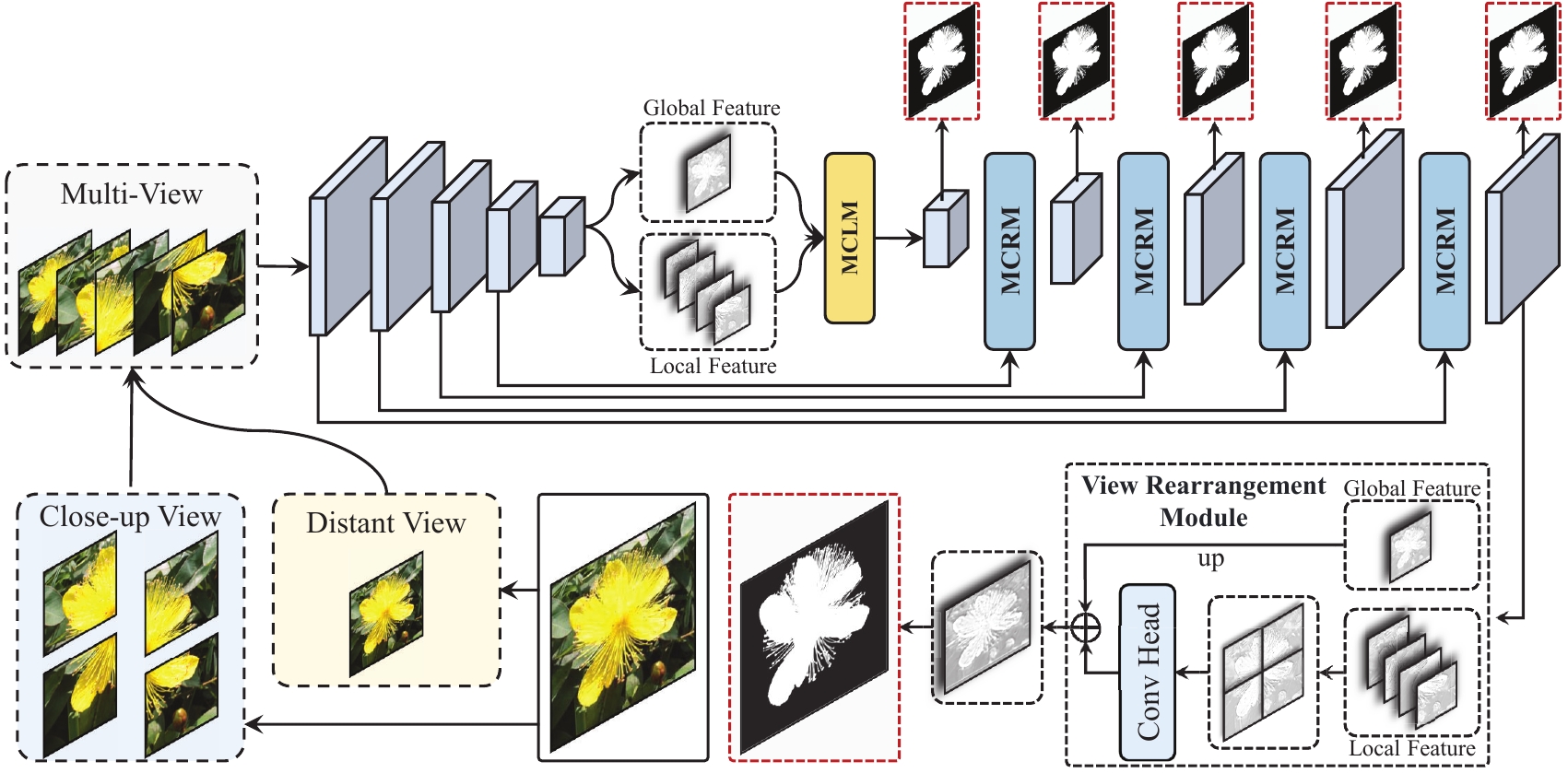}
  \caption{Overall framework of the proposed \mymethod.
    The downsampled original image and non-overlapping local patches are adopted as inputs for the global context and detailed cues, representing distant and close-up views, respectively.
    To enhance object localization and achieve detailed depiction, we propose multi-view complementary localization module (MCLM) and refinement module (MCRM), respectively.
    Besides, a view rearrangement module is introduced to integrate multiple views, thereby generating predictions with highly accurate dominant areas while preserving detailed object structures.
    The \textcolor{red}{red} dashed box indicates the location that is deeply supervised.
  }
  \label{fig:net}
\end{figure*}
\subsection{Multi-view Learning}

Multi-view learning is an emerging direction in machine learning that leverages the use of multiple perspectives to enhance generalization performance~\cite{sun2013survey}.
It involves the utilization of distinct functions to model individual views, and collectively optimizes these functions to exploit other perspectives of the same input data, thereby enhancing overall learning performance~\cite{zhao2017multi}.
In recent years, the integration of multi-view information with deep learning has garnered significant attention in many areas, such as
3D object recognition~\cite{su2015multi, yu2018multi},
3D reconstruction~\cite{wang2021multi, xie2019pix2vox,li2023neuralangelo,wang2023autorecon},
and feature matching~\cite{sattler2016efficient,he2020epipolar}.
Su \etal~\cite{su2015multi} pioneered the utilization of multi-view 2D projected images as inputs, constructing a multi-view convolutional neural network to leverage information from object perspectives for 3D shape recognition. Moreover, Wang \etal~\cite{wang2021multi}proposed a representative multi-view 3D reconstruction scheme, which encodes the relevant information amongst different views to jointly explore multi-level correspondence and associations between the 2D input views and 3D output volume with in a single unified framework.
To this end, we're inspired to split the input with high-resolution image information into multi-view patch sequences to leverage complementary information for a more comprehensive understanding of visual data.

\section{Method}
\label{sec:method}

In this section, we present the proposed approach in detail, including the overall architecture and specific components.

\begin{figure*}[t]
  \begin{subfigure}{0.49\linewidth}
    \centering
    \includegraphics[width=\linewidth]{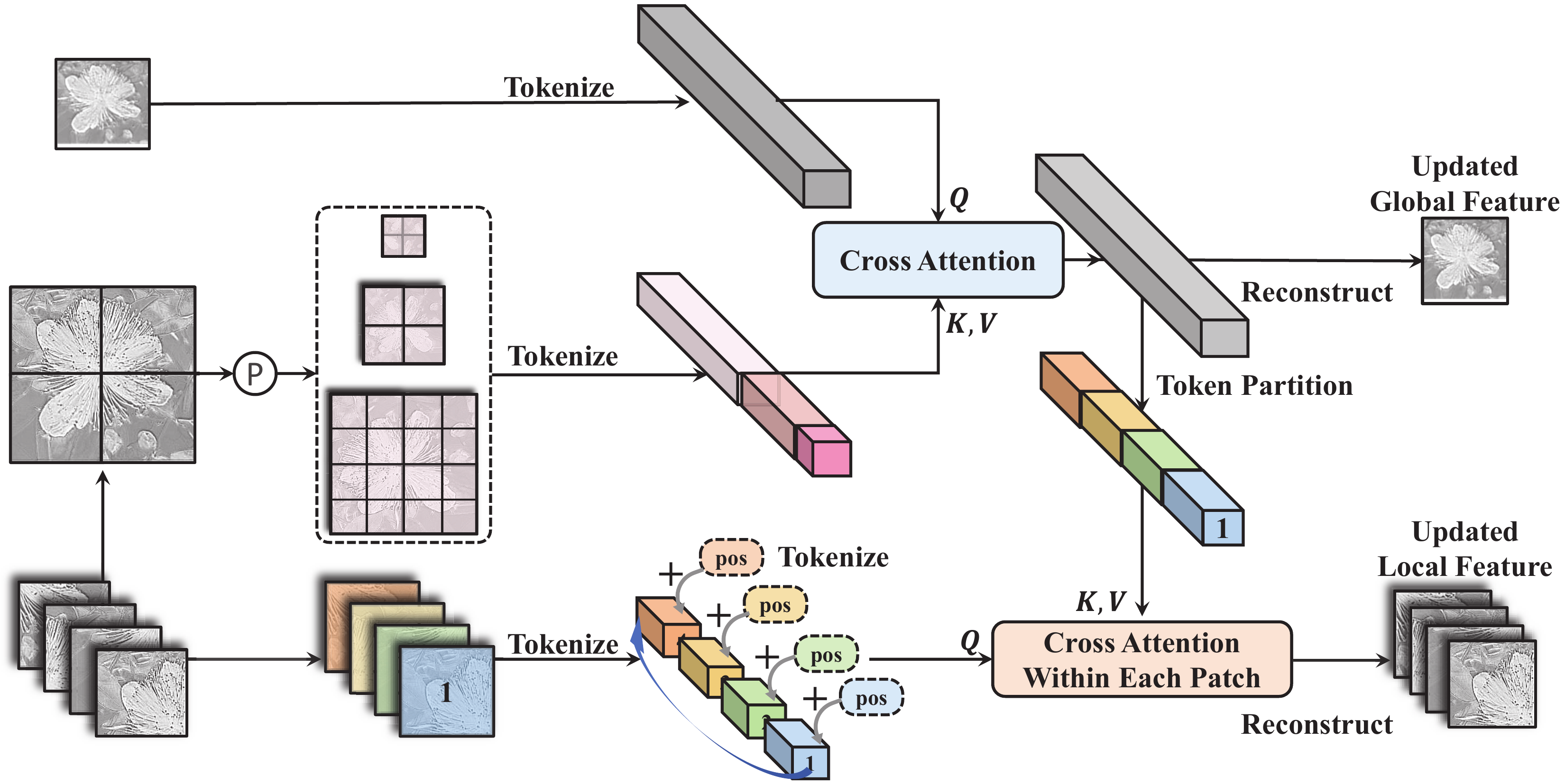}
    \caption{Multi-view Complementary Localization Module.}
    \label{fig:mclm}
  \end{subfigure}
  \hfill
  \begin{subfigure}{0.49\linewidth}
    \centering
    \includegraphics[width=\linewidth]{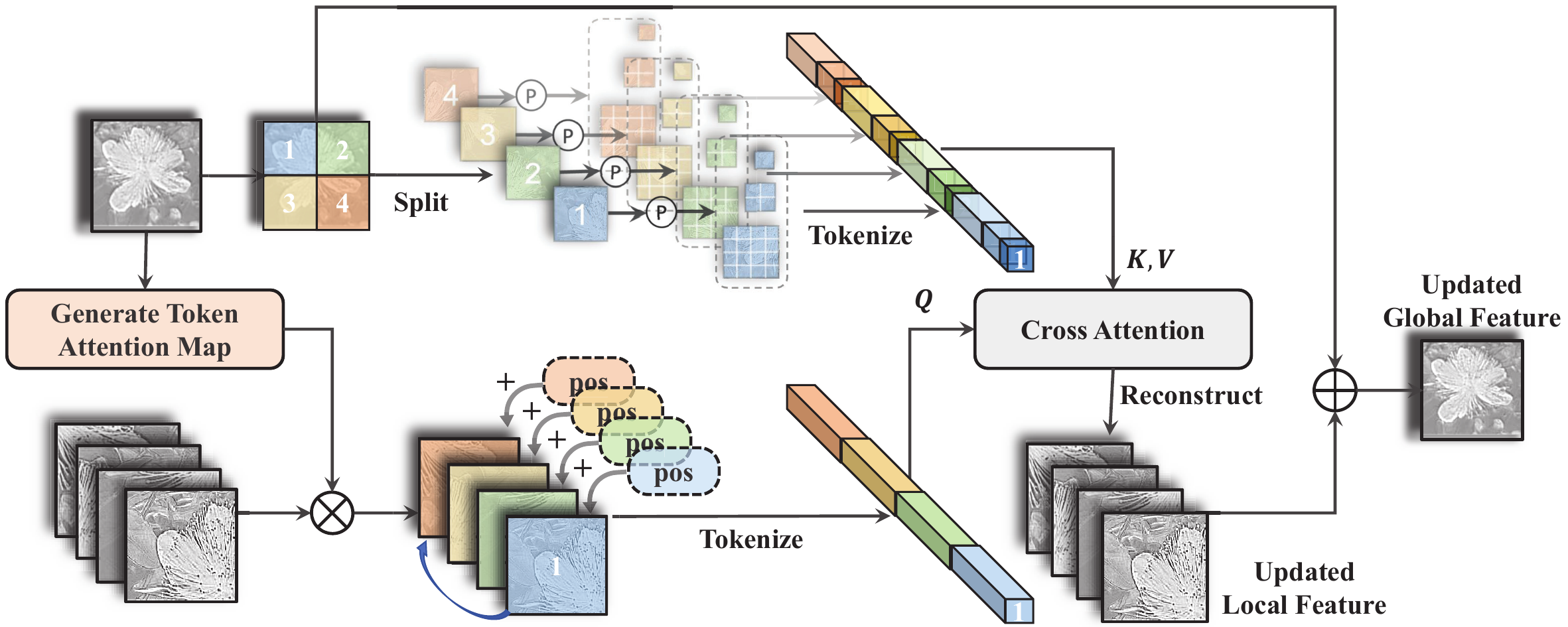}
    \caption{Multi-view Complementary Refinement Module.}
    \label{fig:mcrm}
  \end{subfigure}

  \caption{Pipeline of the proposed multi-view complementary localization and refinement modules.
    \includegraphics[height=1em]{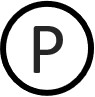} represents the multi-granularity pooling operation.
  }
  \label{fig:module}
\end{figure*}

\subsection{Overall Architecture}

\parhead{Multi-view Input.}
As illustrated in~\cref{fig:net}, the HR image input $\mathbf{I} \in \mathbb{R}^{B \times 3 \times H \times W}$ is resized to create the LR version ${G} \in \mathbb{R}^{B \times 3 \times h \times w}$, which simulates the distant view.
Also, we evenly crop ${I}$ into several non-overlapping local patches $\{{L}_m\}_{m=1}^M \in \mathbb{R}^{B \times 3 \times h \times w}$.
Each of them can be seen as a specific close-up view focusing on the fine-grained texture.
In this paper, we set $M$ to be 4, \ie, $(H,W)=(2h,2w)$, and the corresponding discussion can be found in~\cref{sec:4.4}.

\parhead{Multi-level Feature Extraction.}
${G}$ and $\{{L}_m\}_{m=1}^M$ together make up the multi-view patch sequence, which is fed in batches into the feature extractor to generate the multi-level feature maps, \ie, $\{ {E}_i | {i=1,2,3,4,5} \}$.
Each ${E}_i$ includes representations of both the distant and close-up views.

\parhead{Complementary Localization.}
The feature map ${E}_5$ from the highest level is partitioned along the batch dimension into the two different sets, \ie, global and local features.
They are fed into the multi-view complementary localization module (MCLM) to highlight the positional information about the object within the global representation.
It is subsequently used to guide the local representation for object localization and effectively filter out erroneous information from the close-up view.
After the MCLM, the updated global and local feature maps are concatenated along the batch dimension to form a single feature map $D_5 \in \mathbb{R}^{B \times 3 \times \frac{h}{32} \times \frac{w}{32}}$, which is sent to the well-designed top-bottom decoder.

\parhead{Refinement Decoding.}
Our novel network differs from the classic FPN~\cite{lin2017feature}-like architecture.
We insert the on-the-fly multi-view complementary refinement module (MCRM) in each decoding stage as shown in~\cref{fig:net}.
These models can dynamically optimize missing fine-grained details in the global representation with information from the local representations.
And shallow features are also absorbed layer by layer into the upsampling path in the decoder.

\parhead{Multi-view Integration.}
As illustrated in the bottom-right section of \cref{fig:net}, we introduce a simple view rearrangement module to merge the positional and semantic information from the distant view with the detailed information from the close-up view, into a unified whole.
After the aforementioned steps, we can obtain $D_1^{merge} \in \mathbb{R}^{B \times 3 \times \frac{h}{2} \times \frac{w}{2}}$, whose shape is a quarter of the shape of the original image when $M=4$.
Instead of directly upsampling it by 4 times, we incorporate shallow features\cite{liang2021swinir} as low-level visual cues to further enhance the quality of image segmentation.

\subsection{Multi-view Complementary Localization}
\label{sec:3.2}

To tackle the challenge of jointly localizing objects through distant view and close-up views, we propose the multi-view complementary localization module (MCLM).
The well-constructed process facilitates the proposed model in attaining the holistic scene understanding and effectively identifying potential areas of interest, thereby accomplishing the goal of jointly localizing close-up views and distant views.
First, we divide the $E_5$ into global feature $E_5^G\in \mathbb{R}^{B \times C \times \frac{H}{32} \times \frac{W}{32}}$ and local features $\{E_5^{L_m}\}_{m=1}^M$ where $E_5^{L_m}\in \mathbb{R}^{B \times C \times \frac{h}{32} \times \frac{w}{32}}$ and $M$ denotes the number of local features. 
Subsequently, the local features are assembled into a unified global feature $E_5^{L_g} \in \mathbb{R}^{B \times C \times \frac{H}{32} \times \frac{W}{32}}$ by aligning with their respective positions in the original image. 
To simultaneously obtain the rich visual representation and capture important contextual feature cues, we embed the multi-granularity pooling operation into the vanilla transformer block~\cite{wu2022p2t,zhu2019asymmetric}, which reduces the computational cost of $\mathtt{MHCA}$ by 56.25\% while facilitating deeper interaction between multiple views.
Specifically, we apply multiple average pooling layers with various receptive fields onto the aforementioned unified global feature, thereby generating pyramid feature maps:
\begin{equation}
  \begin{aligned}
    P_1 & = \mathtt{AvgPool}_1(E_5^{L_g}), \\
    P_2 & = \mathtt{AvgPool}_2(E_5^{L_g}), \\
        & \dots,                           \\
    P_n & = \mathtt{AvgPool}_n(E_5^{L_g}),
  \end{aligned}
\end{equation}
where $n$ denotes the number of parallel pooling branches.
And we set the respective receptive fields to be 4, 8, 16 in practice.
These maps are then tokenized and concatenated to be $K$ and $V$ for the $\mathtt{MHCA}$ block:
\begin{align}
  K, V = [T(P_1), T(P_2), \dots, T(P_n)] W^{K,V},
\end{align}
where $W^{K,V} \in \mathbb{R}^{C \times 2C}$ is used to transform all branches.
And $T( \cdot )$ indicates the tokenization operation which is achieved with a flattening process for simplicity.
The operation $[ \cdot ]$ concatenates all sequences into a single one.
Besides, the global feature is also tokenized directly to be $Q$ for $\mathtt{MHCA}$:
\begin{align}
  Q = T(E_5^G) W^Q,
\end{align}
where $W^{Q} \in \mathbb{R}^{C \times C}$ is a projection matrix.
%
Then, $Q$, $K$ and $V$ are fed into $\mathtt{MHCA}$, followed by $\mathtt{LN}$ and $\mathtt{FFN}$ as shown in the ``Cross-Attention'' part of~\cref{fig:mclm}:
\begin{align}
  T^G & = T(E_5^G) + \mathtt{LN}(\mathtt{MHCA}(Q, K, V)), \\
  T^G & = T^G + \mathtt{LN}(\mathtt{FFN}(T^G)).
\end{align}
The updated global token $T^G \in \mathbb{R}^{\frac{HW}{32^2} \times B \times C}$ can be utilized to reconstruct and update the global feature, where we can obtain $F^{G'}$ for further processing.
Besides, in order to utilize it to further assist in the activation of object-related cues in the local field of view, we also rearrange and partition $T^G \in \mathbb{R}^{\frac{HW}{32^2} \times B \times C}$ according to the order of patch tokens, referred as $\{T^{G_m}\}_{m=1}^M$.
To effectively remain the positional correlation between different views, we supplement the position encoding into these local features as shown in~\cref{fig:mclm}.
Subsequently, we tokenize them and apply $\mathtt{MHCA}$ within each patch, where the local tokens is used as $Q$ and the rearranged global tokens as $K$ and $V$:
\begin{align}
  Q_m      & = T(E_5^{L_m}) W^{Q_m},                 \\
  T^{L_m'} & = \mathtt{MHCA}(Q_m, T^{G_m}, T^{G_m}).
\end{align}
Finally, based on the updated local tokens $\{T^{L_m'}\}_{m=1}^M$, a straightforward unflatten and reshape procedure is applied to generate the reconstructed local features $\{E_5^{L_m'}\}_{m=1}^M$, which are then simply concatenated in batches with the updated global feature $E_5^{G'}$ to form the feature map $D_5$ for subsequent processing.

\subsection{Multi-view Complementary Refinement}
\label{sec:3.3}

After the LR global feature provides a broader context aiding in coarse-level identification, we introduce the multi-view complementary refinement module (MCRM) as shown in~\cref{fig:mcrm}.
In this module, local features provide localized and detailed views to enhance the accuracy and robustness of segmentation.
To be specific, the input feature is denoted as \(D_i\), where \(i \in \{1,2,3,4,5\}\) represents the layer number of the decoder.
Similar to the MCLM, we partition the feature \(D_i\) into global feature \(D_i^G\) and local features $\{D_i^{L_m}\}_{m=1}^M$ along the batch dimension.
To filter out background noise from the local features, a one-channel token attention map $A$ is initially generated using a $1 \times 1$ convolution layer followed by a $\mathtt{sigmoid}$ function.
$A$ is subsequently utilized to the modulate feature map and eliminate the background noise, thereby obtaining a purer representation for the object segmentation.
And the aforementioned operations can be formulated as:
\begin{align}
  A                     & = \mathtt{sigmoid}(\mathtt{conv}(D_i^G)),                            \\
  \{D_i^{L_m}\}_{m=1}^M & = \mathtt{split}(A \odot \mathtt{assemble}( \{D_i^{L_m}\}_{m=1}^M)),
\end{align}
where \(\odot\) is the Hadamard product.
$\mathtt{assemble}$ and $\mathtt{split}$ are a pair of opposite operations.
The former rearranges the independent patches into to the original image form, while the latter reverses the process to the patch sequence.
After that, as in the MCLM, the individual position encoding is also added to each local feature to model their positional relationships.
We then tokenize and concatenate these features to serve as $Q$ for the cross attention:
\begin{align}
  T^{L_m}_i & = [T(D_i^{L_1}), T(D_i^{L_2}), \dots, T(D_i^{L_m})], \\
  Q_i       & = [T_i^{L_1}, T_i^{L_2}, \dots, T_i^{L_m}] W^{Q_i}.
\end{align}
Besides, we partition the global feature into corresponding regions based on the original positions of each local feature:
\begin{align}
  \{D_i^{G_m}\}_{m=1}^M = \mathtt{split}(D_i^G).
\end{align}
And then, a similar multi-granularity pooling process as in the MCLM is imposed in these patch-wise features $\{D_i^{G_m}\}_{m=1}^M$ as shown in~\cref{fig:mcrm}, which involves the extraction of contextual information through the utilization of multiple branches with varying receptive fields.
After the transformation and concatenation, we can obtain the multi-sensory tokens $T^{G_m}_i$ with different contextual abstraction levels in the \(m^{th}\) patch.
These tokens are then concatenated into a unified whole, as $K$ and $V$ for the cross attention:
\begin{align}
  K_i, V_i = [T^{G_1}_i , T^{G_2}_i ,\dots, T^{G_m}_i] W^{K_i, V_i}.
\end{align}
During the cross attention operation, we employ a vanilla transformer block to facilitate interaction between informative local tokens and multi-sensory tokens from corresponding regions in the global context.
Then, we reconstruct the updated local tokens to the local features $\{D_i^{L_m'}\}_{m=1}^M$ by adjusting the shape, which are then integrated into the original global feature by the addition operation to obtain globally optimized features $D_i^{G'}$ with enhanced details.
Finally, these two sets of features are concatenated along the batch dimension, resulting in a detail-enhanced feature map:
\begin{align}
  D_i' = [\{D_i^{L_m'}\}_{m=1}^M, D_i^{G'}].
\end{align}
After repetitively stacking the decoding components while continuously integrating the multi-level features from the encoder, we can obtain output features \(D_1'\) with a higher resolution, which incorporates the broader context and the fine-grained locality.

\begin{table*}[t]
  \centering
  \resizebox{0.9\linewidth}{!}{
    \begin{tabular}{cc|ccccccc|ccccc}
\toprule[2pt]
\multirow{3}{*}{\emph{Datasets}}              & \multirow{3}{*}{\emph{Metric}}                        & F\(^3\)Net & GCPANet   & PFNet    & BSANet   & ISDNet    & IFA     &  PGNet      & IS-Net     &FP-DIS& UDUN      &InSPyReNet & {Ours}    
\\ 
        &                       & \cite{wei2020f3net} & \cite{chen2020global}    & \cite{mei2021camouflaged}      & \cite{zhu2022can}     & \cite{guo2022isdnet}     & \cite{hu2022learning}    &   \cite{xie2022pyramid}      & \cite{qin2022highly}      &\cite{zhoudichotomous}&  \cite{pei2023unite}       &  \cite{kim2022revisiting} &   
\\ 
      
                                    &                        & R-50       & R-50       & R-50       & R2-50       & R-50       & R-50    &  S+R   & -           &R-50& R-50       &Swin-B      & Swin-B      \\ \hline
                                    
    \multirow{5}{*}{\emph{\rotatebox{90}{DIS-TE1}}}& \(F_{\beta}^{max}\) \(\uparrow\) & 0.726      & 0.741      & 0.740      &            0.683 &            0.717&            0.673&  0.754&             0.740&0.784&            0.784&0.845& \textbf{0.893}      \\ 
     & \(F^{\omega}_{\beta}\) \(\uparrow\)      & 0.655      & 0.676      & 0.665      &            0.545&            0.643&            0.573&  0.680&             0.662 &0.713&            0.720&    0.788  & \textbf{0.823}      \\ 
                                    & \(E^m_{\phi}\) \(\uparrow\)      & 0.820      & 0.834      & 0.830      &            0.773&            0.824&            0.785&  0.848&             0.820&0.860&            0.864&0.874      & \textbf{0.911}      \\ 
                                    & \(S_m\) \(\uparrow\)             & 0.783      & 0.797      & 0.791      &            0.754&            0.782&            0.746&  0.800&             0.787&0.821&            0.817&0.873& \textbf{0.879}      \\ 
                                    & $\mathcal{M}$   \(\downarrow\)              & 0.074      & 0.070      & 0.075      &            0.098&            0.077&            0.088&  0.067&             0.074&0.060&            0.059&0.043      & \textbf{0.037}      \\ \hline
\multirow{5}{*}{\emph{\rotatebox{90}{DIS-TE2}}}& \(F_{\beta}^{max}\) \(\uparrow\) & 0.789      & 0.799      & 0.796      &            0.752&            0.783&            0.758&  0.807&             0.799&0.827&            0.829&0.894& \textbf{0.925}      \\ 
 & \(F^{\omega}_{\beta}\) \(\uparrow\)      & 0.719      & 0.741      & 0.729      &            0.628&            0.714&            0.666&  0.743&             0.728 &0.767&            0.768&   0.846   & \textbf{0.874}      \\ 
                                    & \(E^m_{\phi}\) \(\uparrow\)      & 0.860      & 0.874      & 0.866      &            0.815&            0.865&            0.835&  0.880&             0.858&0.893&            0.886&0.916& \textbf{0.944}      \\
                                    & \(S_m\) \(\uparrow\)             & 0.814      & 0.830      & 0.821      &            0.794&            0.817&            0.793&  0.833&             0.823&0.845&            0.843&0.905& \textbf{0.915}\\ 
                                    & $\mathcal{M}$  \(\downarrow\)              & 0.075      & 0.068      & 0.073      &            0.098&            0.072&            0.085&             0.065&             0.070&0.059&            0.058&0.036& \textbf{0.030 }     \\ \hline
           \multirow{5}{*}{\emph{\rotatebox{90}{DIS-TE3}}}                         & \(F_{\beta}^{max}\) \(\uparrow\) & 0.824      & 0.844      & 0.835      &            0.783&            0.817&            0.797&             0.843&             0.830&0.868&            0.865&0.919& \textbf{0.936}      \\ 
           & \(F^{\omega}_{\beta}\) \(\uparrow\)      & 0.762      & 0.789      & 0.771      &            0.660&            0.747&            0.705&  0.785&             0.758 &0.811&            0.809&      0.871& \textbf{0.890}      \\ 
                                    & \(E^m_{\phi}\) \(\uparrow\)      & 0.892      & 0.909      & 0.901      &            0.840&            0.893&            0.861&             0.911&             0.883&0.922&            0.917&0.940& \textbf{0.954 }     \\ 
                                    & \(S_m\) \(\uparrow\)             & 0.841      & 0.855      & 0.847      &            0.814&            0.834&            0.815&             0.844&             0.836&0.871&            0.865&0.918& \textbf{0.920 }     \\
            & $\mathcal{M}$  \(\downarrow\)              & 0.063      & 0.068      & 0.062      &            0.090&            0.065&            0.077&             0.056&             0.064&0.049&            0.050&0.034& \textbf{0.031}      \\ \hline
             \multirow{5}{*}{\emph{\rotatebox{90}{DIS-TE4}}}                       & \(F_{\beta}^{max}\) \(\uparrow\) & 0.815      & 0.831      & 0.816      &            0.757&            0.794&            0.790&             0.831&             0.827&0.846&            0.846&0.905& \textbf{0.911  }    \\
             & \(F^{\omega}_{\beta}\) \(\uparrow\)      & 0.753      & 0.776      & 0.755      &            0.640&            0.725&            0.700&  0.774&             0.753 &0.788&            0.792&    0.848  & \textbf{0.857}      \\ 
                                    & \(E^m_{\phi}\) \(\uparrow\)      & 0.883      & 0.898      & 0.885      &           0.815 &  0.873          & 0.847           &             0.899&             0.870&0.906&            0.901&0.936& \textbf{0.944  }    \\ 
                                    & \(S_m\) \(\uparrow\)             & 0.826      & 0.841      & 0.831      &        0.794    &      0.815             &      0.841&       0.811     &             0.830&0.852&            0.849&\textbf{0.905}&0.903      \\ 
            & $\mathcal{M}$  \(\downarrow\)              & 0.070      & 0.064      & 0.072      &    0.107        &         0.079   &         0.085  &             0.065 &             0.072&0.061&            0.059&0.042& \textbf{0.041 }     \\ \hline\hline
\multirow{5}{*}{\emph{\rotatebox{90}{Overall}}}& \(F_{\beta}^{max}\) \(\uparrow\) & 0.789& 0.804& 0.797&0.744 &0.778 &0.755&  0.809 &  0.799&0.831& 0.831&0.891&\textbf{0.916}\\
& \(F^{\omega}_{\beta}\) \(\uparrow\)      & 0.722      & 0.746      & 0.730      &            0.618&            0.707&            0.661&  0.746&             0.726 &0.770&            0.772&   0.838   & \textbf{0.855}      \\ 
& \(E^m_{\phi}\) \(\uparrow\)      & 0.864&0.879 &0.871 &0.811 &0.864 &0.832&  0.885 &  0.858&0.895& 0.892&0.917&\textbf{0.938}\\
& \(S_m\) \(\uparrow\)             &0.816 & 0.831& 0.823&0.789 & 0.812& 0.791&  0.830& 0.819 &0.847& 0.844&0.900&\textbf{0.905}\\
& $\mathcal{M}$   \(\downarrow\)              & 0.071& 0.065& 0.071&0.098 & 0.073& 0.084&  0.063&  0.070&0.057& 0.057&0.039&\textbf{0.035}\\
\bottomrule[2pt]
\end{tabular}
  }
  \caption{Quantitative comparison of DIS5K with 11 representative methods.
    $\downarrow$ represents the lower value is better, while $\uparrow$ represents the higher value is better.
    The best score is highlighted in \textbf{bold}.
    R-50, R2-50, and SwinB respectively denote the utilization of ResNet-50\cite{resnet}, Res2Net-50\cite{res2net}, and Swin-B\cite{liu2021swin} as backbones, while S+R represents the combination of Swin-B and ResNet-50 as a new backbone.
  }
  \label{tab:1}
\end{table*}

\subsection{View Rearrangement}
\label{sec:3.4}

The patch-based local enhancement strategy can retain sufficient texture details for the model, but also introduces the problem of misalignment between neighboring patch boundaries when reorganizing the patches into the image.
In our decoder embedded with the refinement module, the iterative dense interactions between patches and global features alleviate this problem.
And we make further optimizations in the cascaded view rearrangement module as shown in~\cref{fig:net}
Specifically, we split the local features in $D_1'$ along batch dimension and $\mathtt{assemble}$ them into a global form.
To address aforementioned issue, we introduce a convolutional head that consists of three convolutional layers interspersed with $\mathtt{BN}$ and $\mathtt{ReLU}$ layers, to smooth the features as in~\cref{fig:net}.
This architecture is purposefully tailored to prioritize the enhancement of patch alignment.
Subsequently, the aligned feature is added to the global feature split from \(D_1'\) to further enhance image quality, and used to generate the final segmentation map.

\subsection{Loss Function}
As illustrated in the \cref{fig:net}, we incorporate supervision at each layer output of the decoder and the final prediction.
Specifically, the former consists of three components: \(l_{l}\), \(l_{g}\) and \(l_{a}\) for the assembled local representation, the global representation, and the token attention map in the refinement module, respectively.
Note that the side outputs here each require a separate convolutional layer to obtain a single-channel prediction.
And the latter is represented as $l_{f}$.
These components employ the combination of the binary cross-entropy (BCE) loss and the weighted IoU loss, following the common practice in most segmentation tasks~\cite{zhao2022self,pei2023unite,zhoudichotomous}:
\begin{align}
  l = l_{BCE}+l_{IoU}.
\end{align}
To this end, our total loss can be written as:
\begin{align}
  L = l_{f} + \sum_{i-1}^{5} {(l_{l}^i +\lambda_g l_{g}^i+\lambda_a l_{a}^i)},
\end{align}
where \(\lambda_g\) and \(\lambda_h\) are set to 0.3 in our experiment.

\begin{figure*}[t]
  \centering
  \includegraphics[width=0.9\linewidth]{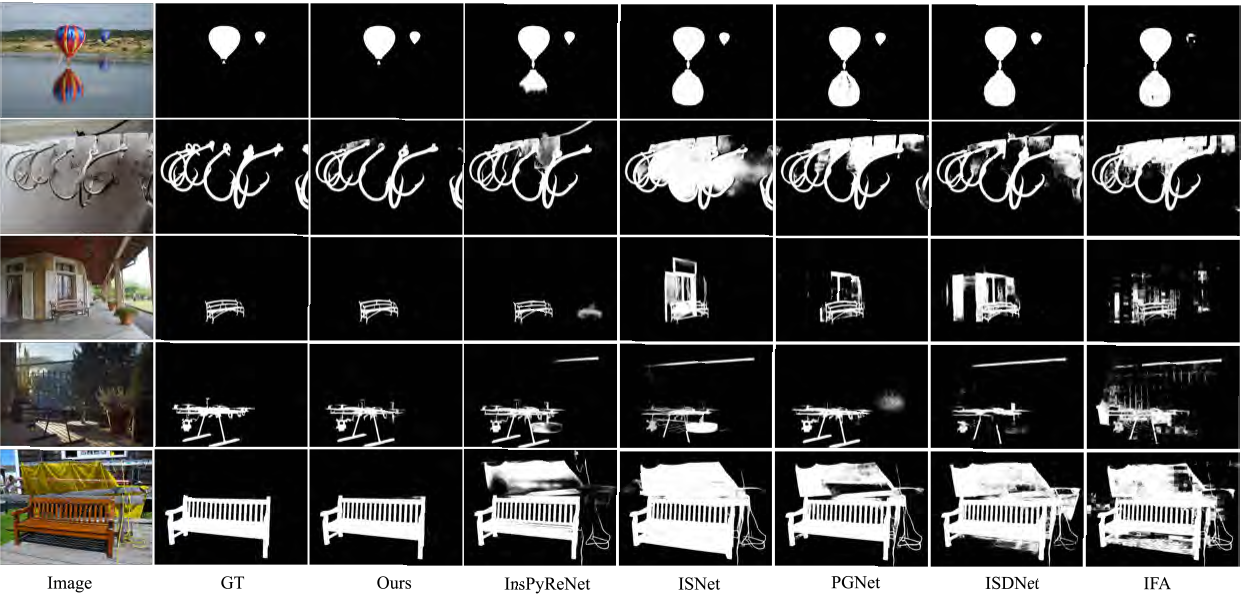}
  \caption{Visual comparison of different DIS methods.}
  \label{fig:5}
\end{figure*}

\section{Experiments}

\subsection{Datasets and Evaluation Metrics}

\parhead{Data Settings.}
We conduct extensive experiments on the DIS5K~\cite{qin2022highly} benchmark dataset, comprising $5,470$ HR images (e.g., 2K, 4K or larger) across $225$ categories. The dataset is partitioned into three subsets: DIS-TR, DIS-VD, and DIS-TE. DIS-TR and DIS-VD consist of $3,000$ training images and $470$ validation images, respectively.
DIS-TE is further divided into four subsets (DIS-TE1, 2, 3, 4) with increasing shape complexities, each containing $500$ images.
With its diverse objects featuring varying geometric structures and appearances, the DIS5K dataset presents higher resolution images, intricate structural details, and superior annotation accuracy compared to existing object segmentation datasets.
As a result, segmentation on DIS5K proves to be challenging and necessitates models with robust capabilities in identifying structural details.

\parhead{Evaluation Metrics.}
For evaluating the results, we adopt some widely used metrics,  including max F-measure (\(F_{\beta}^{max}\))~\cite{perazzi2012saliency},
weighted F-measure (\(F_{\beta}^\omega\)) \cite{weightedFmeasure},
structural similarity measure (\(S_m\))~\cite{fan2017structure},
E-measure (\(E^m_{\phi}\))~\cite{fan2018enhanced}
and mean absolute error (MAE, $\mathcal{M}$)~\cite{perazzi2012saliency}.
\(F_{\beta}^{max}\)  and  \(F_{\beta}^\omega\)  are the maximum and weighted scores of the precision and recall, respectively, where \(\beta^2\) is set to 0.3. \(S_m\) simultaneously evaluates region-aware and object-aware structural similarity between the prediction and mask. \(E^m_{\phi}\) is widely used for evaluating pixel-level and image-level matching. MAE measures the average error of the prediction maps.

\subsection{Implementation Details}
Experiments are implemented in PyTorch on a single RTX 3090 GPU.
During the training phase, the original images are first resized to $1024 \times 1024$. Then, both the With the number of patches set to 4, the resulting patch size is $512 \times 512$.
Consequently, the low-resolution (LR) global image is also resized to $512 \times 512$.
Swin-B~\cite{liu2021swin} is used as the backbone with the pre-trained weights on the ImageNet~\cite{deng2009imagenet}, while other parameters are initialized randomly.
To avoid overfitting, we adopt some data augmentation techniques, including random horizontal flipping, cropping and rotating.
We use the Adam optimizer with an initial learning rate of $0.00001$. The batch size is set to $1$, and the maximum number of epochs is set to $80$.

\subsection{Comparison with State-of-the-arts}
\noindent\textbf{Quantitative Evaluation.}
In Tab.~\ref{tab:1}, we compare our proposed \mymethod with other 11 well-known task-related models, including
F\(^3\)Net~\cite{wei2020f3net},
GCPANet~\cite{chen2020global},
PFNet~\cite{mei2021camouflaged},
BSANet~\cite{zhu2022can},
ISDNet~\cite{guo2022isdnet},
IFA~\cite{hu2022learning},
IS-Net~\cite{qin2022highly},
FP-DIS~\cite{zhoudichotomous},
UDUN~\cite{pei2023unite},
PGNet~\cite{xie2022pyramid},
InSPyNet~\cite{kim2022revisiting}.
For a fair comparison, we standardize the input size of the comparison models to $1024 \times 1024$.
It can be seen that \mymethod significantly outperforms the other models on all the datasets under different metrics.
In particular, ours outperforms the second-best model (InSPyReNet) with the gain of \(2.5\%\) , \(2.1\%\) , \(0.5\%\) , \(0.4\%\) in terms of the \(F_{\beta}^{max}\),  \(E^m_{\phi}\), \(S_m\) and MAE, respectively.
Besides, we evaluate the inference speed for the InSPyReNet and ours.
Both of them are tested under the same NVIDIA RTX 3090 GPU.
Benefiting from the parsimonious single stream design, \mymethod achieves the $4.6$ FPS over the InSPyReNet with the $2.2$ FPS.
\noindent\textbf{Qualitative Evaluation.}
To demonstrate the highly accurate prediction of our model in an intuitive perspective, we visualize the output of some images selected from the test set. As shown in \cref{fig:5}, our model can capture both the accurate object localization and edge details under different complex scenes. In particular, other methods suffer from interference from the salient yellow gauze and shadows, whereas our model allows for a complete segmentation of the chair and accurate differentiation of the interior for each grille (see the last row).

\begin{table}
  \renewcommand\arraystretch{1.25}
  \centering
  \resizebox{0.9\linewidth}{!}{
    \begin{tabular}{ccc|cccc} 
\toprule[2pt]
      HR-Ori&   LR-Dis &  HR-Clo&  \(F_{\beta}^{max} \uparrow\)&  \(E^m_{\phi} \uparrow\)  &  \(S_m \uparrow\)  & $\mathcal{M}$ \(\downarrow\) \\
      \hline 
     \(\checkmark\)& &&  0.822&  0.869&  0.812& 0.058\\ 
     \cline{1-3}
    &\(\checkmark\) &&  0.815&  0.858&  0.801& 0.058\\ 
     \cline{1-3}
     & &\(\checkmark\)&  0.801&  0.814&  0.759& 0.069\\ 
     \cline{1-3}
     \(\checkmark\)& \(\checkmark\)&&  0.875&  0.897&  0.862& 0.041\\  
     \cline{1-3}
     & \(\checkmark\)&\(\checkmark\)&  \textbf{0.893}& \textbf{0.911}& \textbf{0.879}& \textbf{0.037}\\ 
\bottomrule[2pt]
\end{tabular}

  }
  \caption{Ablation  experiments of diverse views inputs. "HR-Ori" is the  high-resolution original images. "LR-Dis" is the low-resolution global images generated by resizing the "HR-Ori".  "HR-Clo" is the 4 non-overlapping patches evenly cropped from the "HR-Ori".}
  \label{tab:3}
\end{table}

\subsection{Ablation Study}\label{sec:4.4}
In this section, we analyze the effects of each component. All the results are tested on the DIS-TE1.

\noindent\textbf{Diverse Views Inputs.}
To investigate the effectiveness of our multi-view input strategy, we conduct a series of experiments involving different inputs, as shown in~\cref{tab:3}.
First, we separately list the in-dependent results of the "HR-Ori", "LR-Dis", "HR-Clo" models. The "HR-Orr" will as the performance anchor to show the effectiveness of the  multi-view inputs strategy. Next, the gap between the first and fourth rows show the effectiveness of low-resolution global images in providing the distant view. Then, the gap between the fourth and last rows can illustrate the necessity of the local patches in providing the close-up view.  Finally, the combined global and local multi-view inputs provide a complete set of target perceptual cues with no mutually exclusive effects on each other.

\noindent\textbf{Effectiveness of MCLM.}
In \cref{tab:2}, the comparison of the second row with last row shows the effectiveness of the proposed  multi-view complementary localization module (MCLM).  The utilization of pyramid pooling for generating multi-sensory tokens enables the identification of targets within tokens at minimal cost and facilitates long-range visual interactions across multiple local views. Therefore, the MCLM guarantees an increase in performance while decreasing speed by only $1.09$ FPS.

\noindent\textbf{Effectiveness of MCRM.}
The gap between the third row and last row verify the effectiveness of the proposed multi-view complementary refinement module (MCRM). Although the pooling operation is used in MCRM, it is only embed in the global features. Once applying cross-attention on shallow features, the computational cost will increases significantly due to the larger feature sizes. By generating multi-sensory tokens, we can reduce the sequence length and thereby greatly reduce the computational burden.

\noindent\textbf{Effectiveness of VRM.}
Thanks to the sufficient feature fusion among local patch features in the decoder, we only need to perform a simple convolution operation at the tail of the MVANet to accomplish an effective view rearrange and obtain a complete high-resolution prediction. Finally, we can see that the combination of MCLM, MCRM and VRM can separately achieve more than 8\%, 4\%, 8\%, 36\% performance gain in terms of the \(F_{\beta}^{max}\),  \(E^m_{\phi}\), \(S_m\) and MAE, compared to the FPN baseline (the first row) with only the original view image input.

\begin{table}
  \renewcommand\arraystretch{1.25}
  \centering
  \resizebox{0.9\linewidth}{!}{
    \begin{tabular}{ccc|ccccc} 
\toprule[2pt]
         MCLM&  MCRM&  VRM&  \(F_{\beta}^{max} \uparrow\)& \(E^m_{\phi} \uparrow\)  &\(S_m \uparrow\)  &$\mathcal{M}$\(\downarrow\) &FPS \(\uparrow\) \\ \hline 
          &  &  &  0.822&  0.869& 0.812& 0.058&\textbf{9.2}\\ 
         \cline{1-3}
         &  \(\checkmark\)&  \(\checkmark\) &  0.880&  0.889& 0.866& 0.038&5.71\\
         \cline{1-3}
         \(\checkmark\)&  &  \(\checkmark\) &  0.884&  0.903& 0.870& 0.041&5.38\\ 
         \cline{1-3}
         \(\checkmark\)&  \(\checkmark\)&  &  0.888&  0.897& 0.874& 0.039&4.76\\
 \cline{1-3}
 \(\checkmark\)& \(\checkmark\)& \(\checkmark\)& \textbf{0.893}& \textbf{0.911}& \textbf{0.879}& \textbf{0.037}&4.62\\ 
\bottomrule[2pt]
\end{tabular}
  }
  \caption{Ablation  experiments  of  each component.}
  \label{tab:2}
\end{table}

\noindent\textbf{Patch Quantity.}
To thoroughly investigate the impact of patch quantity on our work, we crop the original image into 4, 9, and 16 patches, serving as the close-up view inputs. As shown in~\cref{tab:4}, performance degrades as the number of patches increases.
It may be attributed to two reasons: 1) While an increase in patch quantity and a decrease in resolution may enhance processing speed, it also results in reduced information within each close-up view, weakened connectivity between patches, and even instances where diminished receptive fields lead to noise being mistaken for foreground objects. 2) As the resolution decreases for each patch, the resolution of the LR global image correspondingly diminishes, leading to significant information loss that is detrimental to accurate object localization.

\begin{table}
  \renewcommand\arraystretch{1.1}
  \centering
  \resizebox{0.6\linewidth}{!}{
    \begin{tabular}{c|cccc} 
\toprule[2pt]
     Number&  \(F_{\beta}^{max} \uparrow\)&  \(E^m_{\phi} \uparrow\)  &  \(S_m \uparrow\)  & $\mathcal{M}$ \(\downarrow\) \\ \hline 
     4&  \textbf{0.893}&  \textbf{0.911}& \textbf{ 0.879}&\textbf{0.037}\\  
     9&  0.821&  0.856&  0.799&0.058\\ 
     16&  0.717&  0.751&  0.745&0.081\\ 
\bottomrule[2pt]
\end{tabular}

  }
  \caption{Ablation experiments of the number of the patches in the sequence for the close-up view.}
  \label{tab:4}
\end{table}

\section{Conclusion}
In this paper, we tackle the high-accuracy DIS by modeling it as a multi-view object perception problem and provide a parsimonious, streamlined multi-view aggregation network, aiming at making a better trade-off among model designs, accuracy, and the inference speed.
To address the target alignment problem for multiple views, we propose the multi-view complementary localization module to jointly calculate the co-attention region of the target.
Besides, the proposed multi-view complementary refinement module are embed into each decoder block to fully integrate complementary local information and mitigate the semantic deficit of a single view patch, thus the final view rearrangement can be accomplished with only a single convolutional layer.
Extensive experiments show that our model performs well on the DIS dataset.

\noindent\textbf{Acknowledgements.}
This work was supported by the National Natural Science Foundation of China under Grant 62276046 and by Dalian Science and Technology Innovation Foundation under Grant 2023JJ12GX015.

  {
    \small
    \bibliographystyle{ieeenat_fullname}
    \bibliography{main}
  }


\end{document}